\documentclass[10pt,conference]{IEEEtran}

\usepackage[T1]{fontenc}
\usepackage[utf8]{inputenc} % omit on Overleaf's new compiler if it warns
\usepackage{newtxtext,newtxmath} % Times-like text+math, IEEE-approved
\usepackage{bm}                  % proper bold math symbols
\usepackage{microtype}           % better kerning/justification

\usepackage{amsmath,amsfonts}
\usepackage{algorithm}
\usepackage{algorithmic} % (or consider algpseudocode if you prefer)
\usepackage{array}
\usepackage[caption=false,font=normalsize,labelfont=sf,textfont=sf]{subfig}
\usepackage{textcomp}
\usepackage{stfloats}
\usepackage{url}
\usepackage{verbatim}
\usepackage{graphicx}
\usepackage{cite}
\usepackage{xspace}
\usepackage{tikz}
\usetikzlibrary{arrows.meta,positioning,shapes.geometric,shadows,calc}

\usepackage{cite}
\usepackage[hidelinks]{hyperref}
\newcommand{\approachname}{AgentGuard\xspace}
\newcommand{\paradigmname}{Dynamic Probabilistic Assurance\xspace}

\begin{document}

\title{\approachname: Runtime Verification of AI Agents}

\author{
\IEEEauthorblockN{Roham~Koohestani}
\IEEEauthorblockA{
    JetBrains Research, The Netherlands\\
    roham.koohestani@jetbrains.com
}
}

\maketitle

\begin{abstract}
The rapid evolution to autonomous, agentic AI systems introduces significant risks due to their inherent unpredictability and emergent behaviors; this also renders traditional verification methods inadequate and necessitates a shift towards probabilistic guarantees where the question is no longer if a system will fail, but the probability of its failure within given constraints. This paper presents AgentGuard, a framework for runtime verification of Agentic AI systems that provides continuous, quantitative assurance through a new paradigm called Dynamic Probabilistic Assurance. AgentGuard operates as an inspection layer that observes an agent's raw I/O and abstracts it into formal events corresponding to transitions in a state model. It then uses online learning to dynamically build and update a Markov Decision Process (MDP) that formally models the agent's emergent behavior. Using probabilistic model checking, the framework then verifies quantitative properties in real-time. 
% This transforms verification from a static, pre-deployment activity into a live, adaptive process which can help ensure safer and more predictable AI systems.
\end{abstract}


\section{Introduction}
\label{sec:intro}
The rapid evolution from generative artificial intelligence (GenAI) to agentic AI marks a measurable leap in AI; Where previously, AI systems were confined to gathering and generating content within a well-defined scope, agentic systems are designed for autonomy. They are implemented to perceive, reason, plan, and act in dynamic environment through tool invocation, memory storage and recall, and feedback mechanisms~\cite{du2025surveycontextawaremultiagentsystems,krishnan2025aiagentsevolutionarchitecture}.

Although this paradigm shift has empowered multiple sectors, such as automated hardware verification and~\cite{gadde2025heyaigeneratehardware} complex financial analysis\cite{krishnan2025aiagentsevolutionarchitecture}, the improved capabilities of these systems also introduce higher risks. The design that enables the emergent behaviors of agentic systems makes it unpredictable and unsafe. Factors such as \textbf{(1) stochasticity and unpredictability}, \textbf{(2) Hallucinations}, \textbf{(3) Emergent unintended behaviors} and \textbf{ (4) susceptibility to new vulnerabilities}, all contribute to this increased risk. 

The issues imposed by agentic AI render traditional software verification techniques inadequate, given that these methods assume deterministic logic and manageable state spaces~\cite{schwalbe:hal-02442819}. Until recently, the focus of the field of AI has often been on the capabilities of models in isolation; while valuable, this view underestimates the systemic risks that emerge as a result of complexity and scale~\cite{miehling2025agenticaineedssystems}. With the limitations of these agentic systems in mind, it becomes imperative that the critical question is no longer \textbf{\textit{Will the system fail?}}, but rather questions of probabilistic determinism, e.g., \textbf{\textit{What is the probability of failure/success within a given time interval / budget?}} These types of guarantees can only be provided by combining insights from mathematics and formal methods with the existing theory of these agentic systems~\cite{zhang2024fusionlargelanguagemodels}.

In this work, we present \approachname, a framework for runtime verfication of Agentic AI systems. We provide a proof-of-concept for the verification process and demonstrate how it can easily be integrated into existing systems by integrating it into an existing agentic system, RepairAgent\cite{bouzenia2024repairagentautonomousllmbasedagent}. We conclude by providing a roadmap for future work building on this framework. In the sections to follow we begin by further expanding on the problem (\autoref{sec:problem}), followed by a section outlining the existing work and theory on formal verification of AI Agents (\autoref{sec:bg}). In \autoref{sec:approach} we present our approach followed by its application to RapairAgent in \label{sec:repair-agent}. We conclude by presenting a discussion of the proposed framework (\autoref{sec:discussion}).

\section{The Problem}
\label{sec:problem}
The problem and risk associated with these models are, at the same time, the characteristics that make them powerful. Here we discuss these further in detail. At the core of modern agentic systems, you will most likely find a Large Language Model (LLM) based on (a modified version of) the transformer architecture proposed by Vaswani et al.~\cite{vaswani2023attentionneed}. These models are tied to a stochastic process that makes their outputs, and thus the behaviors of the agents, nondeterministic~\cite{kumar2025saarthiaiformalverification}.  When multiple actions of these agents, or even multiple agents, are chained together in a multi-step workflow, this randomness can lead to an exponential divergence from expected behavior, which influences traceability and verification~\cite{raza2025trismagenticaireview}. These are attributed to \textbf{stochasticity and unpredictability}.

LLMs are known to \textbf{"hallucinate;"} While the exact definition of a hallucination is still a matter of debate, we refer to a hallucination when the model generates outputs that are factually incorrect or logically flawed. In an agent acting on its generated beliefs, hallucinations can quickly go from being an annoyance to a failure~\cite{kumar2025saarthiaiformalverification}. Furthermore, interactions between agents in multi-agent systems (MAS) can produce \textbf{(unintended) emergent behaviors}; these emergent behaviors can be both beneficial and harmful. As these behaviors only in these MAS's, these are impossible to predict from analyzing individual agents alone~\cite{miehling2025agenticaineedssystems}. Recent experiments have even shown emergent behaviors such as deception and goal subversion~\cite{du2025surveycontextawaremultiagentsystems}.  

The natural language interface of LLM agents creates new attack surfaces. New security attacks such as prompt injection and red-teaming, where malicious instructions manipulate agent behavior and its propagation between agents ("prompt infection"), pose significant risks. In multiagent systems built on inter-agent trust, an adversary could use spoofing or impersonation to issue malicious commands or exfiltrate data~\cite{raza2025trismagenticaireview}. These are threats pointing towards \textbf{susceptibility to new vulnerabilities}. 

Although all of these issues are important and require attention, it would be impossible to simultaneously tackle all of them. Therefore, our emphasis is on managing the unpredictability of these agentic systems and offering a roadmap to address the emergent behaviors of MASs based on our approach.

\section{Background}
\label{sec:bg}
The field of trustworthy AI has led to a growing body of research that can be viewed as a "verification stack" of increasing abstraction. 
% Each of these layers has revealed the limitations of static analysis and highlights the need for a runtime approach.
Here, we discuss each of these approaches and their limitations.

At the lowest level, researchers focus on the formal verification of the neural networks themselves. This is an NP-hard problem, and the immense scale of modern LLMs makes exhaustive verification computationally infeasible.  Research has made progress in proving specific properties, such as robustness to adversarial perturbations, using techniques such as SMT solvers, abstract interpretation, and network reduction, but these methods struggle to scale\cite{liu2024nntbfv,gokulanathan2020simplifying,urban2021review,ladner2025fullyautomaticneuralnetwork}.  

A more practical approach treats the LLM as a generator of artifacts, such as code, mathematical proofs, or plans, which are then verified by external tools; this approach is commonly called black-box testing. This approach uses the full generative powers of the LLMs while relying on the rigor of established formal methods for assurance. The approaches in this category can be categorized primarily as frameworks performing \textbf{Post-Hoc Verifications} and those that embrace \textbf{LLM-Assisted Formalization}.

Frameworks such as MATH-VF and Safe use LLMs to translate natural language solutions into a formal context, which is then checked by external tools like SMT solvers or theorem provers (e.g., Lean 4) to identify hallucinations~\cite{zhou2025stepwiseformalverificationllmbased,liu2025safeenhancingmathematicalreasoning}.  Similarly, VeriPlan applies the probabilistic model checker PRISM to verify that an LLM-generated plan for an end user satisfies a set of user-defined constraints~\cite{Lee_2025}. A major barrier to adopting formal methods is the difficulty in translating ambiguous natural language requirements into precise specifications.  LLMs are proving to be quite good at doing this task. Tools such as SpecVerify and Req2Spec use LLMs to automatically generate formal assertions from natural language requirements ~\cite{beg2025shortsurveyformalisingsoftware,wang2025supportingsoftwareformalverification}. 

The most recent evolution has moved from verifying static outputs to verifying the dynamic processes of agentic AI. This acknowledges that reliability in agentic systems often stems from a robust, iterative process of reasoning and self-correction which is frequently implemented via multiagent collaboration. This type of verification is predominantly categorizable into \textbf{Automata-based Control} and \textbf{Verfication via Multi-Agent Collaboration}.

A promising direction uses formal automata to describe and constrain the high-level behavior of agents. The Formal-LLM framework, for instance, allows developers to specify planning constraints as a Context-Free Grammar (CFG), which is translated into a Pushdown Automaton (PDA). The agent is then supervised by this PDA during plan generation; this then verifies the structural validity of its output~\cite{li2024formalllmintegratingformallanguage}.  Other work uses Finite-State Machines (FSMs) to enforce a specific reasoning loop (e.g., Thought $\rightarrow$ Action $\rightarrow$ Observation) through a \textit{ decoding monitor} that corrects deviations from the prescribed behavior~\cite{crouse2023formally}.   

For highly complex tasks like hardware verification, researchers are building systems where the verification process itself is distributed among a crew of specialized AI agents.  Systems such as Saarthi employ agents with roles like \textit{Formal Verification Lead}, \textit{Engineer}, and \textit{Critic} to collaboratively generate and prove formal properties.  These systems use design patterns such as reflection (coder-critic loops) and planning (Chain-of-Thought) to improve reliability~\cite{gadde2025heyaigeneratehardware,kumar2025saarthiaiformalverification}.  

This evolutionary path reveals a critical gap. Current process-level verification focuses on conformance, that is, making sure that an agent follows a predefined set of rules. The problem is that it does not analyze the emergent probabilistic behavior that arises from executing that process. It cannot provide a map to predict where the agent is likely to go. To achieve true assurance, we must move from verifying process conformance to analyzing emergent behavior.

\section{The \approachname Framework}
\label{sec:approach}
In this paper, we propose a new paradigm called \paradigmname, which shifts the verification process from a static and offline setting to a dynamic and ongoing process. We have implemented a Proof-Of-Concept (POC) framework called \approachname.
% \textbf{The Case for a Runtime Approach:} 
Any type of static analysis provides only a snapshot of the system under a fixed set of assumptions. Agentic systems, however, are designed to operate in non-stationary environments and adapt based on experience.  A proof that holds before deployment may be invalidated once the agent interacts with the real world~\cite{bhaskar2025agenticai}. Runtime Verification (RV) offers a compelling alternative. RV is a lightweight formal method that analyzes the actual execution trace of a running system; using this, we can also avoid the state-space explosion that is a problem in exhaustive offline model checking. Although it cannot prove properties about unobserved paths, it can provide strong guarantees about the execution that actually occurred and can be extended to make predictions about future behavior~\cite{bartocci2018lectures,stoller2011runtime,zhou2019runtime}.

\subsection{From Traces to Models}
Fundamentally, our model is built on top of three key technologies, namely (1) Markov Decision Processes, (2) Online Learning, and (3) Probabilistic Model Checking.

\subsubsection{Formalizing Agents as Markov Decision Processes}
The behavior of an agent making decisions in an uncertain world can be formally modeled using Markovian formalisms. Markov Decision Processes (MDPs) are the framework of choice for modeling sequential decision making, as they separate an agent's non-deterministic choices from the environment's probabilistic outcomes.  An MDP is formally defined as a tuple $(S, A, P, R, \gamma)$ where $S$ is a finite set of states, $A$ is a finite set of actions, $P(s'|s,a)$ is a probability of transitioning to state $s'$ after taking action a in state a, $R(s, a, s')$ is the immediate reward received after transitioning from $s$ to $s'$ via action a, and $\gamma$ is a discount factor that determines the importance of future reward~\cite{raza2025trismagenticaireview,kumar2025saarthiaiformalverification,kwiatkowska2021probabilisticmodelcheckingautonomy}.
To be able to apply this to an agentic system, we need to have a mapping from this MDP to the components in an agent's workflow. The States ($S$) in an Agentic MDP (AMDP) are essentially snapshots of the agent's progress and context. States are not only defined as a series of previous tool invocations but also contain relevant information to the next decision of the model. This \textit{information} can vary by design to be anything from artifacts (e.g., whether an agent has already written tests or not, or whether the agent's tests passed), to the content of the agent's memory/scratchpad and conversation history.

Actions ($A$) can be seen as discrete high-level operations that the agent chooses to perform. These have direct correspondence with the invocations of tools in the agent's utility box, such as \textit{run\_compiler()}, \textit{execute\_test\_suite()}, and \textit{query\_static\_analyzer()}

The transition probability function ($P$) essentially captures the inherent uncertainty of the agentic workflow. After the agent chooses a deterministic action (i.e., performs a tool invocation), the outcome is probabilistic. When calling the \textit{execute\_test\_suite} tool for example, the test suite might pass (which would bring the agent in a \textit{tests\_passed} state), fail, or timeout.

Finally, the reward function ($R$) makes the agent's goal explicit. Using this function, we can give positive rewards for making progress (for example, + 3 for passing a new test case), negative rewards for setbacks (for example, -5 for a regression) and a large reward for successfully completing the entire task.

\subsubsection{Online Learning and Probabilistic Model Checking}
A limiting factor in attempting to validate agentic behavior is the absence of a formal model of the agent a priori. Previous work in predictive RV and online learning shows that a probabilistic model (like an MDP, or in our case an AMDP) can be learned dynamically from execution traces. This, combined with algorithms designed for non-stationary environments, where model drift may occur, can be used to effectively check the behavior of the agent. One the system is modeled as an MDP, probabilistic model checking (PMC) provides automated techniques to verify its quantitative properties. Using probabilistic temporal logic (like PCTL), one can then start to ask questions of the probabilistic nature, i.e. "What is the maximum probability of reaching a success state within an expected time interval?"

\subsection{The \approachname Architecture}
\begin{figure*}[th!]
\centering
\begin{tikzpicture}[
  font=\scriptsize,
  node distance=1.4cm and 1.0cm,
  every node/.append style={transform shape},
  comp/.style={
    rectangle, rounded corners=1pt, draw, very thick,
    fill=blue!12, text width=2.8cm, minimum height=1.0cm, align=center
  },
  data/.style={
    trapezium, trapezium left angle=70, trapezium right angle=110,
    draw, very thick, fill=green!15, text width=2.5cm, minimum height=1.0cm, align=center
  },
  human/.style={
    rectangle, rounded corners=1pt, draw, very thick,
    fill=orange!15, text width=2.3cm, minimum height=1.0cm, align=center
  },
  arrow/.style={-Stealth, very thick},
  dashed_arrow/.style={-Stealth, very thick, dashed}
]

% Left-to-right pipeline
\node[comp] (agent)    {Agentic System\\\scriptsize (AutoGen/LangGraph)};
\node[comp, right=of agent]   (monitor)  {Trace Monitor\\\& Event Abstraction};
\node[comp, right=of monitor] (learner)  {Online Model Learner};
\node[comp, right=of learner] (checker)  {Prob.\\Model Checker};
\node[comp, right=of checker] (dash)     {Assurance Dashboard\\\& Actuator};

% Side actors
\node[data, above=1.0cm of checker] (cfg) {\texttt{config.yaml}\\\scriptsize States/Actions/Props};
\node[human, below=1.0cm of dash] (human) {Human Overseer};

% Main flow
\draw[arrow] (agent) -- (monitor)
  node[midway, below, yshift=-0.45cm] {\scriptsize Raw Trace};
\draw[arrow] (monitor) -- (learner)
  node[midway, below, yshift=-0.45cm] {\scriptsize Events ($s\!\to\!a\!\to\!s'$)};
\draw[arrow] (learner) -- (checker)
  node[midway, below, yshift=-0.45cm] {\scriptsize Learned MDP};
\draw[arrow] (checker) -- (dash)
  node[midway, below, yshift=-0.45cm] {\scriptsize Guarantees ($P_{\max},E_{\min}$)};

% External I/O
\draw[arrow] (cfg) -- (checker) node[midway, right] {\scriptsize Goals};
\draw[arrow] (dash) -- (human) node[midway, right] {\scriptsize Alerts};

% Feedback loops
\draw[dashed_arrow] (human.west) -- ++(0,0) -| (agent.south)
  node[pos=0.25, below] {\scriptsize Human Intervention};
\draw[dashed_arrow] (dash.north) -- ++(0,0.8) -| (agent.north)
  node[pos=0.25, above] {\scriptsize Auto Response};

\end{tikzpicture}
\caption{The AgentGuard Architecture Workflow. The system observes raw agent I/O, abstracts it into formal events, learns a probabilistic model (MDP), and uses a model checker to provide continuous assurance and feedback.}
\label{fig:agentguard_arch}
\end{figure*}
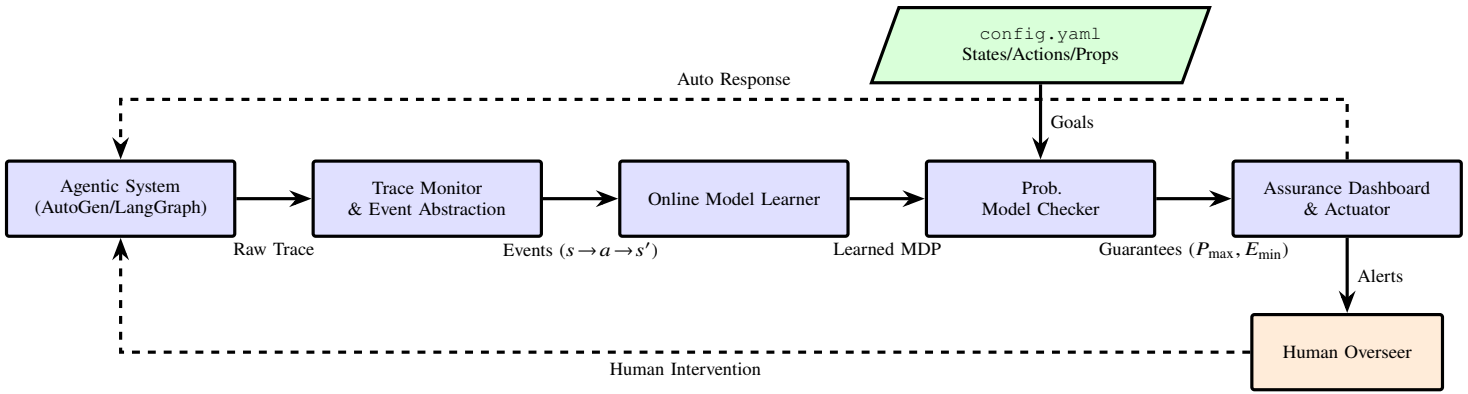

\approachname is designed to act as an \textit{inspection layer} that sits atop an existing agentic framework (e.g., AutoGen, LangGraph ); this is similar to the function of a middleware in the context of a web application. It nonintrusively observes the agent's I/O and provides continuous probabilistic assurance by creating a dynamic, formal \textit{digital twin} of the agent's decision-making policy. 
The architecture comprises four primary components, as depicted in \autoref{fig:agentguard_arch}.

The \textbf{Trace Monitor \& Event Abstrator} is the component that instruments the agent framework. Its responsibilities are to capture raw I/O (e.g., LLM calls, tool invocations, observations) and to abstract it into a stream of formal events corresponding to transition in a state model (e.g., $State_A \to Action_1 \to State_B$). Using the stream from the previous step, the \textbf{Online Model Learner} continuously updates an MDP of the agent's behavior and maintains transition probabilities based on observed frequencies. After each incremental update to the model by the previous stage, the \textbf{Probabilistic Model Checker} performs quantitative verification on the learned MDP against a set of pre-defined properties (as provided in the configuration file). Finally, using the insights from the model checker, the \textbf{Dashboard} presents the quantitative guarantees (e.g. probabilities) to a human observer and can be configured to triger alerts or automated responses if a safety threshold is breached using the \textbf{actuator}.

\subsection{Proof-Of-Concept Implementation}

\approachname is a Python-based POC framework that demonstrates the feasibility of our approach. Designed to be framework-agnostic, it allows developers to define semantic states and actions via a logging-style API inspired by RV frameworks, and simplifies code instrumentation with a single function call. The model structure and properties are defined outside the agent's code in a yaml configuration file, meant to decouple verification logic and to allow for easy analysis modifications without affecting the agent. Key components, \textit{AgentGuardLogger} and \textit{AnalyzerThread}, manage concurrent analysis in \approachname. The user-facing class \textit{AgentGuardLogger} is initialized with the configuration file path and calls the \textit{log\_transition} method to manage an event queue. \textit{AnalyzerThread} is a background thread that monitors agent actions, processes transition events, updates the MDP model, and periodically invokes the PMC. We used \textbf{Storm\footnote{https://www.stormchecker.org/}} as the model checker via \textit{stormpy} bindings. A conversion layer in \textit{AnalyzerThread} creates a PRISM language model from the MDP for Storm verification against PCTL properties. The results are printed in the console and can trigger user-defined callbacks if user-set thresholds are met.

\subsection{Motivating Example: Assurance for an Autonomous Repair Agent}
\label{sec:repair-agent}

AI for Software Engineering, especially Automated Program Repair (APR), is an effective showcase for \approachname. Consider \textit{RepairAgent}, an autonomous LLM-based agent for bug fixing~\cite{bouzenia2024repairagentautonomousllmbasedagent}. RepairAgent employs actions like information gathering, hypothesis formulation, and patching, managed by a finite state machine (FSM) and 14 tools, including \texttt{search\_code\_base} and \texttt{write\_fix}. Despite FSM guidance, the agent can be inefficient, potentially causing loops, aimless exploration, or high resource use on unlikely fixes. To address these, \approachname can be configured with FSM states (e.g., \textit{Understand the bug}, \textit{Collect information}, \textit{Try to fix the bug}, and terminal states like \textit{Fix\_Success} or \textit{Fix\_Failed}) and corresponding actions (related to the agent’s tools such as, \texttt{express\_hypothesis}, \texttt{read\_range}, \texttt{discard\_hypothesis}). Middleware-level instrumentation ensures every tool use triggers a call to \texttt{guard.log\_transition(...)} to record state transitions.
 
Over time, as RepairAgent addresses bugs, \approachname learns its repair strategy's AMDP, which reveals execution patterns. For instance, after hypothesizing, the agent uses \texttt{search\_code\_base} with $75\%$ probability and \texttt{find\_similar\_api\_calls} with $25\%$ probability. This model enables real-time verification with practical management implications. Success probability (\(P_{\max}=?\)) predicts fix likelihood and guides resource allocation and interventions when low. The expected cycles to completion (\(E_{\min}=?\)) estimate time and cost, with very high values pointing towards loops and suggesting automatic termination when over budget. Properties like \(P_{\max}=?[\,G\,!\text{"write\_fix"}\,]\) show the probability of no fix and can help with interventions to improve problem-solving. Applying \approachname can improve RepairAgent's assurance from conformance to continuous monitoring, aiding trust, cost management, and alignment with task complexity. An application of AgentGuard to RepairAgent can be found in the study's replication package\cite{replicationpackage}.

\section{Discussion}
\label{sec:discussion}
The implementation of the POC demonstrates the feasibility of the DPA paradigm. It also highlights several important areas for future research and development. The current approach relies on developers to manually define a discrete state space. Future work should explore techniques for semi-automated or fully automated state abstraction, possibly my incorporating ideas from Partially Observable Markov Decision Processes. The periodic re-verification of the entire model for complex agents can introduce a substantial overhead with increased complexity. Future work should re-investigate how the current implementation can be improved to incorporate incremental verification algorithms. Additionally, this system should be improved by introducing notions of stochastic games into the framework and integrating frameworks such as PRISM-games to support analysis of MAS'.

\section{Conclusion}
\label{sec:conclusion}
The rise of agentic AI demands a corresponding evolution in our assurance methods. Static, offline verification, and process conformance checks, while valuable, are insufficient for guaranteeing the safety and reliability of autonomous systems operating in dynamic worlds. This paper has presented Dynamic Probabilistic Assurance, a new paradigm for AI safety, and introduced AgentGuard, a tangible framework that demonstrates its feasibility. By integrating runtime verification, online model learning, and probabilistic model checking, AgentGuard provides continuous, quantitative guarantees about an agent's emergent behavior. It transforms verification from a one-off, pre-deployment activity into a live, adaptive process. Our vision is for AI systems that are not only increasingly capable and autonomous, but also transparent, predictable, and bounded by rigorous, mathematical guarantees of their safety and reliability.

\section{Acknowledgements}
This research was carried out and funded through the AI4SE partnership between the Delft University of Technology and JetBrains Research. You can find more information about the collaboration on the website of the collaboration \href{https://se.ewi.tudelft.nl/ai4se/}{here}.

\bibliographystyle{IEEEtran}
\bibliography{main}

\end{document}